\documentclass[10pt,twocolumn,letterpaper]{article}

\usepackage{cvpr}
\usepackage{times}
\usepackage{epsfig}
\usepackage{graphicx}
\usepackage{amsmath}
\usepackage{amssymb}
\usepackage{bm}
\usepackage{caption}

\usepackage{color}
\usepackage{cite}
\usepackage{makecell}
\usepackage{booktabs}
\usepackage{multirow}

\usepackage[pagebackref=true,breaklinks=true,letterpaper=true,colorlinks,bookmarks=false]{hyperref}

\cvprfinalcopy % *** Uncomment this line for the final submission

\def\cvprPaperID{5340} % *** Enter the CVPR Paper ID here

\ifcvprfinal\fi
\begin{document}

\title{PPR10K: A Large-Scale Portrait Photo Retouching Dataset with \\ Human-Region Mask and Group-Level Consistency}

\author{\textbf{Jie Liang}$^{1, 2}$\footnotemark[1],\; \textbf{Hui Zeng}$^{1, 2}$\footnotemark[1],\;  \textbf{Miaomiao Cui}$^{2}$,\;  \textbf{Xuansong Xie}$^{2}$ and \textbf{Lei Zhang}$^{1,2}$\footnotemark[2]\\
$^1$The HongKong Polytechnic University,\;  $^2$DAMO Academy, Alibaba Group\\
}

\maketitle

\renewcommand{\thefootnote}{\fnsymbol{footnote}}
\footnotetext[1]{Equal contribution.}
\footnotetext[2]{Corresponding author. This work is supported by the Hong Kong RGC RIF grant (R5001-18).}

\begin{abstract}

Different from general photo retouching tasks, portrait photo retouching (PPR), which aims to enhance the visual quality of a collection of flat-looking portrait photos, has its special and practical requirements such as human-region priority (HRP) and group-level consistency (GLC). HRP requires that more attention should be paid to human regions, while GLC requires that a group of portrait photos should be retouched to a consistent tone. Models trained on existing general photo retouching datasets, however, can hardly meet these requirements of PPR. To facilitate the research on this high-frequency task, we construct a large-scale PPR dataset, namely PPR10K, which is the first of its kind to our best knowledge. PPR10K contains $1, 681$ groups and $11, 161$ high-quality raw portrait photos in total. High-resolution segmentation masks of human regions are provided. Each raw photo is retouched by three experts, while they elaborately adjust each group of photos to have consistent tones. We define a set of objective measures to evaluate the performance of PPR and propose strategies to learn PPR models with good HRP and GLC performance. The constructed PPR10K dataset provides a good benchmark for studying automatic PPR methods, and experiments demonstrate that the proposed learning strategies are effective to improve the retouching performance. Datasets and codes are available: \href{https://github.com/csjliang/PPR10K}{https://github.com/csjliang/PPR10K}.

\end{abstract}

\begin{figure*}[t]
		\centering
		\includegraphics[width=0.95\textwidth]{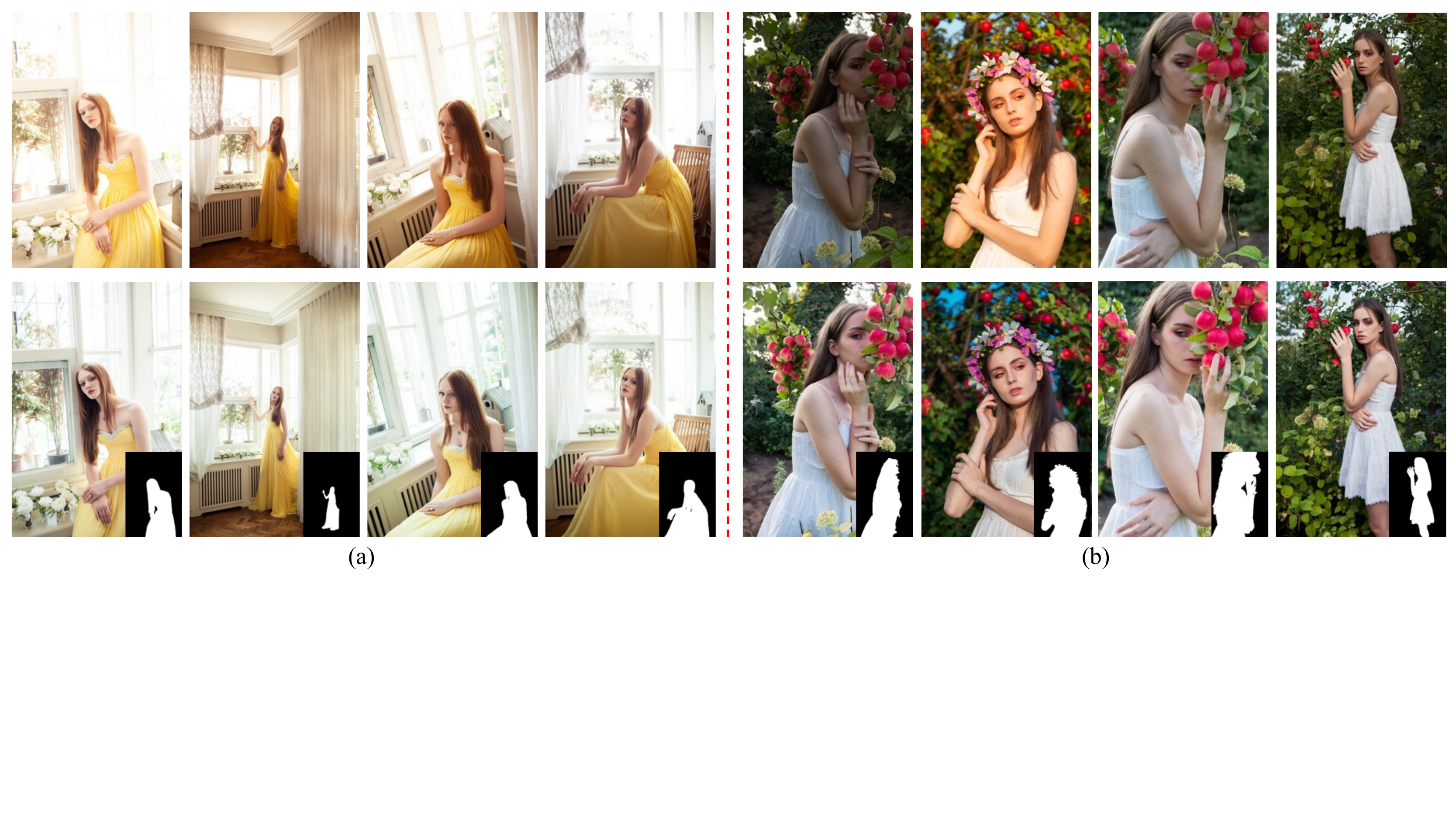}
		\vspace{-8.5em}
		\captionof{figure}{\label{introduction}Examples of a group of photos from the PPR10K dataset (better viewed in color). \textit{Top}: the raw photos; \textit{bottom}: the retouched results from one expert and the human-region masks.
		The raw photos exhibit poor visual quality and large variance in subject views, background contexts, lighting conditions and camera settings. The retouched results demonstrate both good visual quality and group-level consistency.}
\end{figure*}

\section{Introduction}

Portrait photography has a vast range of applications in scenarios such as wedding, birthday, graduation, anniversaries, advertisements, personal recording or creations. To ensure high quality of the finalist photos, photographers tend to capture as many raw photos with high dynamic ranges as possible. However, a set of raw photos might be flat-looking and present inconsistent tone due to the variations of subject view, illumination condition, background contexts and camera settings, as shown in the top row of Figure~\ref{introduction}. A fast retouching on a large set of raw photos is necessary before feeding back to customers for photo selection, followed by fine-grained editing. 

While a set of common standards or styles of pre-retouching are widely accepted and followed, most portrait photos are retouched manually, which is very tedious and time-consuming on the large size and highly redundant raw photos. Automatic portrait photo retouching is thus highly desired as it can save a huge amount of tedious human labor and significantly improves the efficiency of the entire portrait photography pipeline, bringing better experience for both photographers and customers.

Different from general-purpose photo retouching tasks, portrait photo retouching (PPR) has two special and practical requirements: human-region priority (HRP) and group-level consistency (GLC). HRP means that human-related region in a portrait photo should have higher priority and be paid more attention. The first row of Figure~\ref{introduction}(a) shows a set of typical examples, where the backgrounds are over-exposed while the human regions are under-exposed. For such cases, retouching should improve the exposure of human regions while preserving as many details as possible in the backgrounds. GLC requires a group of portrait photos, which are usually taken on the same subject at the same scene but have different subject views, lighting conditions and even camera settings, to be adjusted to a consistent tone, as shown in the bottom row of Figure~\ref{introduction}. 

To the best of our knowledge, existing general-purpose photo retouching or enhancement datasets \cite{bychkovsky2011learning, ignatov2017dslr, hasinoff2016burst, cai2018learning} and models \cite{wang2013naturalness, park2018distort, hu2018exposure,  kosugi2019unpaired, yan2014learning, gharbi2017deep, yan2016automatic, deng2018aesthetic, chen2018deep, he2020conditional, zeng2020learning} do not touch the above two requirements and thus can hardly satisfy the demands of automatic PPR. To facilitate research on this important and high-frequency task, in this paper, we construct the first large-scale PPR dataset which contains $11,161$ (in $1,681$ groups) high-quality raw portrait photos, namely PPR10K dataset hereafter. The raw photos are captured by various DSLR camera devices, covering a wide range of scenes, subjects, lighting conditions and camera settings. Each raw photo is independently adjusted by 3 expert retouchers with rich experience in professional photography studios, resulting in three versions of high-quality retouched targets. Besides the highly informative raw photos and their retouched results, we also provide a high-resolution human-region mask for each photo to make better use of HRP. Each group of photos is elaborately adjusted to ensure GLC. We believe this dataset will provide a valuable benchmark to facilitate the research on automatic PPR.

With the PPR10K dataset, we define a set of objective measures to evaluate the performance of automatic PPR in terms of both HRP and GLC. We also propose corresponding learning strategies to improve the retouching quality of trained PPR models. Specifically, we define human-region weighted measures based on the provided mask, which contribute to achieving better visual quality on subject areas. Explicitly defining the GLC on the image space is very challenging because of the large variations of content in a group of photos. We find that the GLC can be reliably evaluated based on the statistics in CIELAB color space.
We also propose an efficient way to simulate the intra-group variations using individual images, which is proven effective to improve the GLC performance.
Considering that previous general-purpose photo retouching models obtain poor performance on the PPR task, we re-implemented representative state-of-the-art photo retouching and enhancement methods, and report their performance on our dataset for a convenient and fair comparison. 

The contributions of this paper are two-fold.
First, we construct the first large-scale and high-quality PPR dataset with human-region masks and group-level consistent targets, providing a valuable benchmark to facilitate research on this important task.
Second, we propose a set of objective measures and learning strategies to evaluate and optimize the PPR models.
Extensive experiments verified the effectiveness of the proposed dataset, measures and learning strategies both quantitatively and qualitatively.

\section{Related Works}

\begin{figure*}[t]
		\centering
		\includegraphics[width=0.98\textwidth]{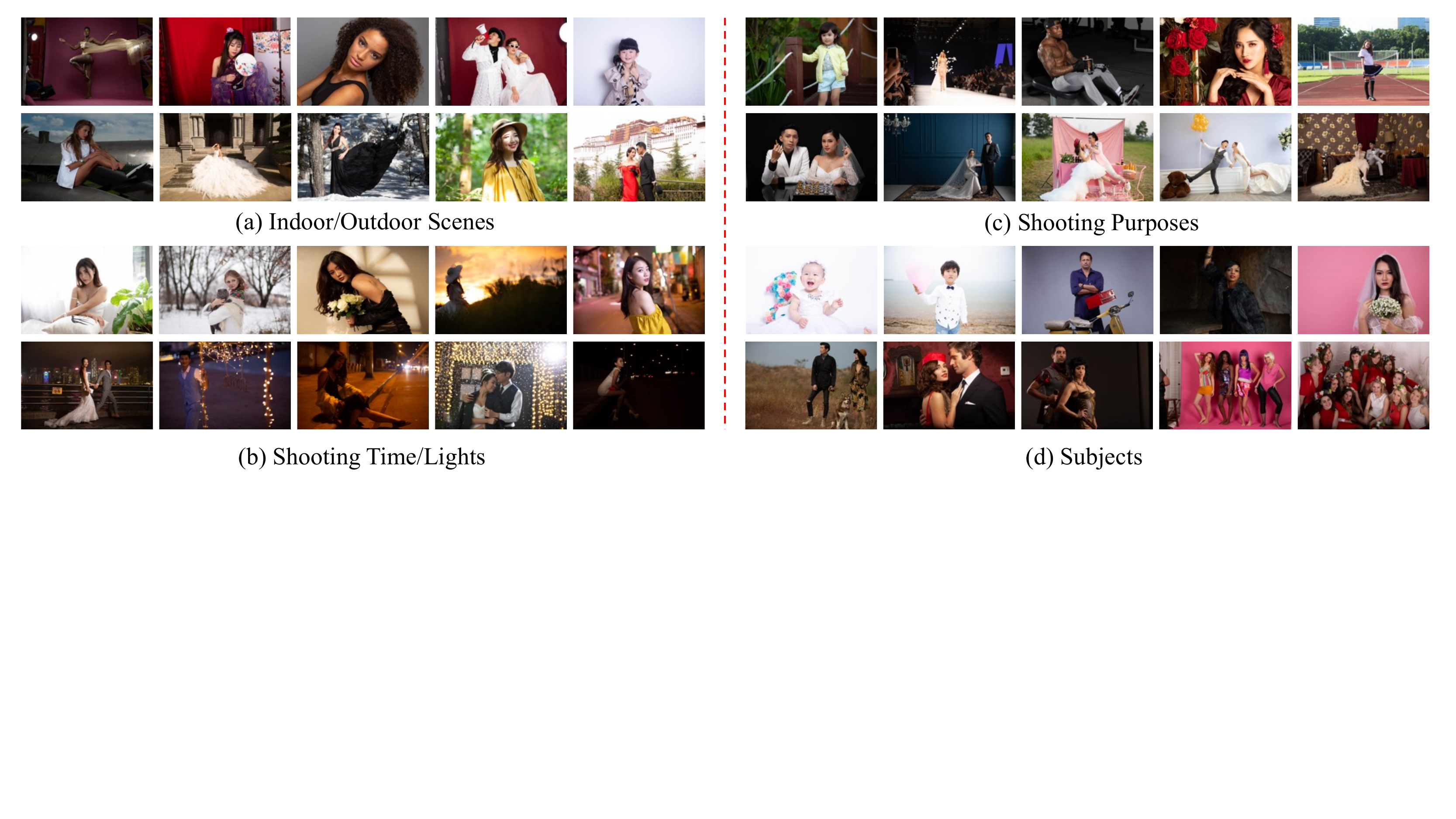}
		\vspace{-11.5em}
		\caption{Visual examples to demonstrate the diversity of the proposed dataset, e.g., different indoor/outdoor scenes, light conditions, shooting purposes and human subjects.
			\label{dataset}}
\end{figure*}

\subsection{Photo Enhancement Datasets}

High-quality datasets are the foundations of learning based photo enhancement or retouching research~\cite{bychkovsky2011learning, ignatov2017dslr, hasinoff2016burst, cai2018learning}.
Bychkovsky \etal~\cite{bychkovsky2011learning} constructed the pioneering FiveK dataset, which contains $5,000$ raw photos of general scenes together with five versions of retouched targets. This dataset has successfully facilitated the research of automatic photo retouching and enhancement \cite{gharbi2017deep, chen2018deep, zeng2020learning}. 
Ignatov~\etal~\cite{ignatov2017dslr} constructed the DPED dataset with an aim to learn a mapping from low-quality photos captured by mobile devices to the counterparts captured by high-end DSLR cameras. This dataset mainly consists of photos in general scenes such as landscapes and street views, and serves as a benchmark for general-purpose photo enhancement task.
There are also datasets focusing on enhancing the dynamic range and contrast of photos~\cite{hasinoff2016burst, cai2018learning}, where the ground-truths are elaborately generated via fusing multiple frames.

Despite the great efforts, the above datasets are constructed on general scenes, where portrait photos only take a minority and receive no special treatment.
In addition, they only consider the visual quality of each individual photo rather than a group of photos that is commonly encountered in portrait photography.
As a result, the models trained on them are unsuitable for the PPR task.
In this paper, we elaborately construct a larger-scale PPR dataset, fulfilling the HRP and GLC requirements of portrait photography.

\subsection{Photo Retouching Methods}
Photo retouching~\cite{hu2018exposure,  kosugi2019unpaired, takahashi1998photographic, mukherjee2008enhancement, cai2018learning, kim1997contrast, gijsenij2011computational, finlayson2004shades, yuan2012automatic, mantiuk2008display} aims to enhance the visual aesthetic quality of an image, which is conventionally achieved via professional tools, \eg, CameraRaw\footnote{https://www.adobe.io/apis/creativecloud/camera-raw.html}, or hand-craft operations like look-up tabels (LUTs)~\cite{karaimer2016software}.
However, these manual tools rely heavily on the empirical knowledge and perceptual aesthetic judgment of the well-trained artists, therefore beyond the abilities of non-professional users.
Some learning-based methods~\cite{bychkovsky2011learning, yan2014learning, jobson1997properties, rahman1996multi, sen2011automatic} based on hand-crafted features have been developed, yet can hardly satisfy the practical demands due to their limited representation capacity against the vast range of image contents and light conditions.

Various deep-learning-based schemes~\cite{park2018distort, yan2016automatic, chen2018deep, deng2018aesthetic, wang2019underexposed, gharbi2017deep} have recently been presented, benefiting from the FiveK dataset~\cite{bychkovsky2011learning} and deep convolutional neural networks~\cite{he2016deep, simonyan2014very}.
Most of these deep models, however, are limited by the input resolutions or the processing time in practice.
For real-time processing on high-resolution images, \eg, images with more than 24M pixels, Gharbi~\etal~\cite{gharbi2017deep} proposed the HDRNet putting most computation on downsampled images.
He~\etal~\cite{he2020conditional} proposed to approximate a sequence of base operations such as brightness or contrast adjustments via a light-weight MLP, while Kosugi~\etal~\cite{kosugi2019unpaired} introduced a reinforcement learning framework to estimate the parameters of these operations.
As a new state-of-the-art, Zeng~\etal~\cite{zeng2020learning} proposed to learn an image-adaptive 3-dimensional LUT (3D LUT), which can retouch 4K images in a speed of more than 500fps with appealing tones.
Nonetheless, the above-mentioned methods do not touch the HRP and GLC requirements, partially due to the lack of training data.
In this paper, based on the constructed dataset, we propose two learning strategies to improve the performance of PPR, providing a benchmark for further research.

\section{The PPR10K Dataset}

As discussed before, existing photo retouching datasets and models cannot fulfill the requirements of PPR. 
To solve these problems, we construct a large-scale and high-quality PPR (PPR10K) dataset.

\noindent\textbf{Challenges: }
To construct a valuable PPR dataset that fulfills the real-world requirements, we have to overcome several challenges.
First, the photos should in raw format with high-quality. However, unlike the abundant and easily available compressed JPG images, raw photos are much more difficult to obtain on the internet.
Second, the dataset should be large-scale and cover a wide range of real cases, in terms of shooting purpose, human subjects, background scenes, lighting conditions as well as the usage of camera devices, which further increases the cost of data collection. 
Third, high-quality retouched results (with both good visual quality and group-level consistency) and human-region masks should be provided to learn effective PPR models. These requirements make the labeling process expensive and cumbersome.

\noindent\textbf{Data Collection and Selection: } 
To obtain as many raw portrait photos as possible, we negotiated with many individual photographers and professional photography studios to purchase raw portrait photos in groups from them for research purposes only. We also purchased from several paid material websites that provide raw format portrait photos. During data collection, we have elaborately control the diversity of raw photos in terms of shooting purpose (e.g., wedding, birthday, graduation, anniversaries, advertisements, personal recording and creation), human subjects (including babies, children, younger, couples and worldwide people), background scenes (including indoor and outdoor, lighting conditions (from day to night, winter to summer), and usage of camera devices (covering a wide series of high-end DSLR cameras of Canon, Nikon and Sony). The diversity of collected photos is shown in Figure~\ref{dataset}.

We initially collected more than $25,000$ raw photos then conducted several rounds of selection. We first discarded photos without human subjects, with low quality such as severe motion blur or de-focus, or containing inappropriate information. We further carefully checked photos group by group, removing outliers (photos with very different content from the group) and duplicated ones (photos with almost the same content). After the screening, we finally obtained a total of $11, 161$ portrait photos in $1, 681$ groups, and each group contains $3\sim18$ photos with same the subjects captured in the same scene at consecutive time. Two typical groups of photos are shown in Figure~\ref{introduction}.

\noindent\textbf{Data Labeling: }
To obtain high-quality ground-truths, we hired $3$ expert retouchers, all of whom have more than $5$ years of experience working in the professional photographic industry, to retouch the raw photos independently, using the CameraRaw in PhotoShop.
Each retoucher was required, based on their own domain knowledge, to retouch the raw photos to satisfy the output standards of professional portrait photography studios with two major requirements. First, each photo should be retouched to visually pleasing to the perception of common people, especially for the human-regions. Second, a group of photos should be adjusted to have a consistent tone. Retouchers were allowed to adjust any operations in CameraRaw without changing content or introducing geometric distortions. In addition, the retouching of each expert was also required to be self-consistent among similar scenes, which is important for learning a stable and robust retouching model. We also hired another expert to double-check the retouched results and conducted several rounds of feedback-and-repair to ensure high-quality of the ground-truths. The retouching style of the three experts is shown in \textbf{supplementary file}.

Considering the high priority of human regions and their complicated illuminations in portrait photos, we also provide human-region masks for learning better retouching models. To save the annotation cost, the masks were first generated using an internal-developed portrait segmentation algorithm which was trained on a set of human matting datasets and supports segmenting photos up to 100 megapixels. We then manually check and refine the failure cases on some difficult scenes such as underwater, extremely low light, glass-reflex and occluded cases.

\begin{figure}[t]
		\centering
		\includegraphics[width=0.44\textwidth]{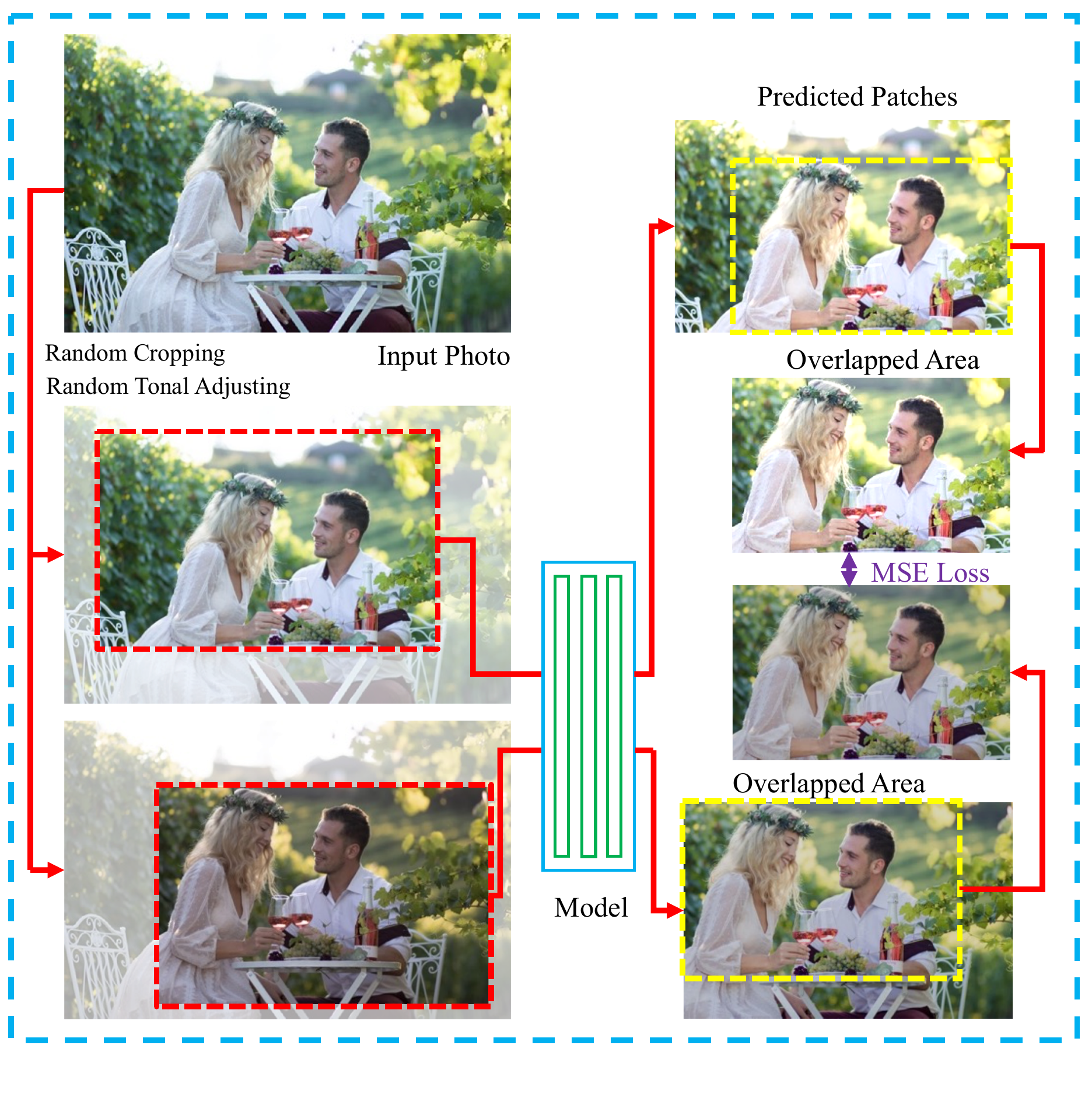}
		\vspace{-1.5em}
		\caption{Illustration of the GLC learning strategy.\label{GLC_pipeline}}
\end{figure}

\noindent\textbf{Discussions: }
Despite the high-quality of our constructed dataset, it leaves several challenges in learning an effective portrait retouching model. First, both the raw photos and human-region masks have very high resolution ranging from 4K to 8K, which requires retouching models to be highly efficient. Second, the diversity of content and lighting conditions in various scenes requires the models to be flexible and content-adaptive. Third, the demand of group-level consistency requires the models to be robust and stable, which is critical for practical applications.

\begin{figure*}[t]
		\centering
		\includegraphics[width=0.96\textwidth]{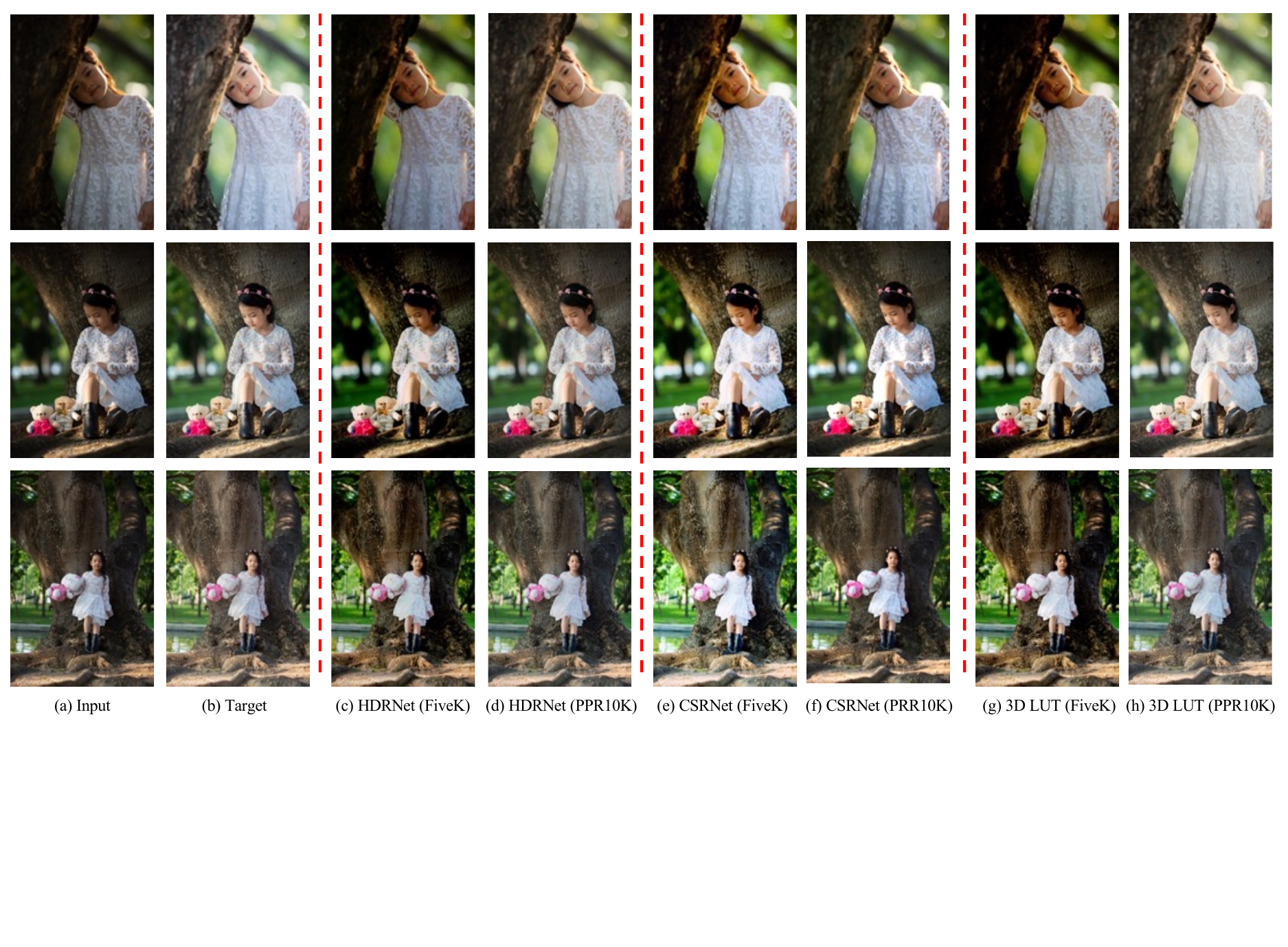}
		\vspace{-8.8em}
		\caption{Visual comparisons between models trained on the FiveK (c, e, g) dataset and the proposed PPR10K (d, f, h) dataset. The targets in (b) are from PPR10K-a.
			\label{visual_comparison}}
\end{figure*}

\section{Measures and Learning Strategies}

Based on the PPR10K dataset, we define a set of measures to quantitatively evaluate the performance of a PPR method. We also propose learning strategies to optimize the HRP and GLC requirements of the PPR task.

\subsection{Basic Measures}

Following the common practice in general-purpose photo enhancement task \cite{bychkovsky2011learning, gharbi2017deep,zeng2020learning}, we first define two basic measures, including the peak signal-to-noise ratio (PSNR) and the CIELAB color difference \cite{wiki:color_difference}. Given an input portrait photo $\bm{I}$, denote by $\hat{\bm{I}}$ and $\bm{Y}$ its predicted version of a PPR model and the target retouched by a human expert, respectively. We can easily obtain their conversions in Lab color space, which are denoted by $\bm{I}^{Lab}$, $\hat{\bm{I}}^{Lab}$ and $\bm{Y}^{Lab}$. Similar to the PSNR defined based on the $L_2$-distance in sRGB color space, the color difference is defined as the $L_2$-distance in CIELAB color space with
$
	\triangle E_{ab} = \|\hat{\bm{I}}^{Lab}-\bm{Y}^{Lab}\|_2.
$
Compared to the sRGB color space, the CIELAB color space is more perceptually uniform and is widely used to tune the tones of photos \cite{wiki:cielab}. 

\subsection{Human-centered Measures}

Considering the higher priority of human regions in portrait photos, we further define two human-centered measures, which can be naturally achieved by putting higher weights to human regions than background regions, leveraging the provided human-region masks in the PPR10K dataset.
Given a photo $\bm{I}$ of $H\times W$ resolution, we can construct its weighting matrix $\bm{W}_{\bm{I}}=[w_{i,j}]\in R^{H \times W}$, where $w_{ij}=1$ for background regions human regions and $w_{ij}=\alpha$ $(\alpha\le 1)$ for background regions. We empirically fix $\alpha=0.5$. The human-centered PSNR ($PSNR^{HC}$) and color difference ($\triangle E_{ab}^{HC}$) can be consequently defined. To save space, we only provide the formula of $\triangle E_{ab}^{HC}$ as:
\begin{equation}
	\triangle E_{ab}^{HC} = \lVert \bm{W}_{\bm{I}}\odot\hat{\bm{I}}^{Lab} - \bm{W}_{\bm{I}}\odot\bm{Y}^{Lab} \rVert_2.
\end{equation}
where $\odot$ denotes the element-wise matrix multiplication.

\subsection{Group-level Consistency Measure}

\label{GLC_section}

Different from the above measures based on individual photos, the group-level consistency (GLC) measures the variations in tone and color among a group of photos. This measure can hardly be defined in the image space since the image contents in a group of photos are not aligned (refer to Figure \ref{introduction}). A reasonable GLC measure should be sensitive to the change of global tone and color appearance, and simultaneously should be robust to the change of image content. It is worth mentioning that the content change is restricted in a group of photos that have the same subject and similar background. Inspired by the practice in white balance \cite{finlayson2004shades}, we define the GLC measure based on the statistics of color components. 

Specifically, given a group of predicted photos $[\hat{\bm{I}}_{1}, \hat{\bm{I}}_{2}, \cdots, \hat{\bm{I}}_{m}]$, we first calculate the mean color components of each photo $[\mu_{\hat{\bm{I}}_{1}}, \mu_{\hat{\bm{I}}_{2}}, \cdots, \mu_{\hat{\bm{I}}_{m}}]$ to represent their global tone and color appearance. The GLC measure is then defined as the variance of mean color components:
\begin{equation}
\begin{aligned}
	\mathcal{M}_{GLC} = \sum_{c}Var(\mu_{\hat{\bm{I}}_{1}^c}, \mu_{\hat{\bm{I}}_{2}^c}, \cdots,  \mu_{\hat{\bm{I}}_{m}^c}),\label{GLC_measure}
\end{aligned}
\end{equation}
where $c$ denotes a color channel which can be chosen from $\{R,G,B,L,a,b\}$ or the combinations of them.
Through extensive quantitative studies, we empirically found that the GLC measure based on the combination of $a$ and $b$ channels is the most suitable and stable choice. The study details can be found in the \textbf{supplementary file}.

\begin{table*}[t]
		\caption{Quantitative comparisons among the baseline and ablation methods. The PPR10K-a/b/c indicate the GTs retouched by three experts. For each measure, the `LR' column reports the results tested on $360$p images and the `HR' column reports the results tested on original resolutions. The $\uparrow$ and $\downarrow$ denote that larger or smaller is better, respectively.}
		\vspace{-1em}
		\footnotesize
		\label{comparison}
		\begin{center}
			\begin{tabular}{p{0.5cm}<{\centering}p{2.7cm}p{1.5cm}p{0.7cm}<{\centering}p{0.7cm}<{\centering}p{0.7cm}<{\centering}p{0.7cm}<{\centering}p{0.7cm}<{\centering}p{0.7cm}<{\centering}p{0.7cm}<{\centering}p{0.7cm}<{\centering}p{0.7cm}<{\centering}p{0.7cm}<{\centering}p{0.7cm}<{\centering}}
				\toprule
				\multirow{2}*{\#}&\multirow{2}*{Method}&\multirow{2}*{Dataset}& \multicolumn{2}{c}{$PSNR \uparrow$} &  \multicolumn{2}{c}{$\triangle E_{ab} \downarrow$} & \multicolumn{2}{c}{$PSNR^{HC} \uparrow$} & \multicolumn{2}{c}{$\triangle E_{ab}^{HC} \downarrow$} & \multicolumn{2}{c}{$\mathcal{M}_{GLC} \downarrow$}\\
				&&&LR&HR&LR&HR&LR&HR&LR&HR&LR&HR \\
				\midrule
				 1&HDRNet~\cite{gharbi2017deep} &PPR10K-a&23.93&23.06&8.70&9.13&27.21&26.58&5.65&5.84&14.83&14.37 \\
				 2&CSRNet~\cite{he2020conditional}  &PPR10K-a&22.72&22.01&9.75&10.20&25.90&25.19&6.33&6.73&12.73&12.66 \\
				 3&3D LUT~\cite{zeng2020learning} &PPR10K-a&25.64&25.15&6.97&7.25&\textbf{28.89}&28.39&4.53&4.71&11.47&11.05 \\
				 4&3D LUT+HRP  &PPR10K-a&\textbf{25.99}&\textbf{25.55}&\textbf{6.76}&\textbf{7.02}&28.29&\textbf{28.83}&\textbf{4.38}&\textbf{4.55}&10.81&10.32 \\
				 5&3D LUT+GLC  &PPR10K-a&25.06&24.39&7.39&7.81&28.34&27.67&4.80&5.06&9.98&9.77 \\
				 6&3D LUT+HRP+GLC  &PPR10K-a&25.31&24.60&7.30&7.75&28.56&27.86&4.75&5.03&\textbf{9.95}&\textbf{9.68} \\
				 
				 \midrule
				 7&HDRNet~\cite{gharbi2017deep} &PPR10K-b&23.96&23.51&8.84&9.13&27.21&26.55&5.74&5.92&13.21&13.04 \\
				 8&CSRNet~\cite{he2020conditional}  &PPR10K-b&23.76&23.29&8.77&9.28&27.01&26.62&5.68&5.90&11.82&11.73 \\
				 9&3D LUT~\cite{zeng2020learning} &PPR10K-b&24.70&24.30&7.71&7.97&27.99&27.59&4.99&5.16&9.90&9.52 \\
				 10&3D LUT+HRP  &PPR10K-b&\textbf{25.06}&\textbf{24.66}&\textbf{7.51}&\textbf{7.73}&\textbf{28.36}&\textbf{27.93}&\textbf{4.85}&\textbf{5.00}&9.87&9.60 \\
				 11&3D LUT+GLC  &PPR10K-b&24.16&23.39&8.15&8.70&27.48&26.71&5.25&5.61&9.17&8.92 \\
				 12&3D LUT+HRP+GLC  &PPR10K-b&24.52&23.81&7.93&8.42&27.82&27.12&5.12&5.44&\textbf{9.01}&\textbf{8.73} \\
				 
				 \midrule
				 13&HDRNet~\cite{gharbi2017deep} &PPR10K-c&24.08&23.66&8.87&9.05&27.32&26.93&5.76&5.99&14.76&14.28 \\
				 14&CSRNet~\cite{he2020conditional}  &PPR10K-c&23.17&22.85&9.45&9.87&26.47&26.09&6.12&6.54&14.64&14.22 \\
				 15&3D LUT~\cite{zeng2020learning} &PPR10K-c&25.18&24.78&7.58&7.85&28.49&28.09&4.92&5.09&13.51&13.16 \\
				 16&3D LUT+HRP  &PPR10K-c&\textbf{25.46}&\textbf{25.05}&\textbf{7.43}&\textbf{7.69}&\textbf{28.80}&\textbf{28.38}&\textbf{4.82}&\textbf{4.98}&13.49&13.06 \\
				 17&3D LUT+GLC  &PPR10K-c&24.53&23.94&8.10&8.49&27.87&27.29&5.25&5.49&12.96&\textbf{12.75} \\
				 18&3D LUT+HRP+GLC  &PPR10K-c&24.59&24.01&8.02&8.39&27.92&27.33&5.20&5.43&\textbf{12.76}&12.79 \\
				\bottomrule

			\end{tabular}
		\end{center}
	\end{table*}

\begin{table}[t]
		\caption{Quantitative results of models trained on the FiveK dataset and evaluated on the PPR10K dataset, where a, b, and c denote the GTs retouched by three experts.}
		\vspace{-1em}
		\footnotesize
		\label{fivek}
		\begin{center}
			\begin{tabular}{p{1cm}p{0.5cm}<{\centering}p{0.9cm}<{\centering}p{0.7cm}<{\centering}p{0.9cm}<{\centering}p{0.9cm}<{\centering}p{0.7cm}<{\centering}}
				\toprule
				%\multirow{2}*{Measure}& \multicolumn
				Method&Dataset&$PSNR$&$\triangle E_{ab} $&$PSNR^{HC} $&$\triangle E_{ab}^{HC} $&$\mathcal{M}_{GLC}$ \\
				\midrule
				HDRNet&a&18.20&17.22&21.44&11.27&20.76\\
				CSRNet&a&19.86&14.07&23.06&9.15&13.97\\
				3D LUT&a&19.92&13.75&23.79&8.90&13.85\\
				HDRNet&b&18.74&16.31&22.00&10.63&20.76\\
				CSRNet&b&19.65&14.47&22.83&9.40&13.97\\
				3D LUT&b&19.74&14.08&23.42&9.22&13.85\\
				HDRNet&c&19.71&14.81&22.96&9.65&20.76\\
				CSRNet&c&19.81&14.57&23.06&9.46&13.97\\
				3D LUT&c&20.03&13.90&23.01&8.85&13.85\\
				\bottomrule
			\end{tabular}
		\end{center}
	\end{table}

\subsection{Learning Strategies}

Optimizing the basic measures and human-centered measures is straightforward. We simply employ the human-region weighted MSE loss on sRGB color images to optimize a model with these measures:
\begin{equation}
	\mathcal{L}_{HC} = \lVert \bm{W}_{\bm{I}}\odot\hat{\bm{I}} - \bm{W}_{\bm{I}}\odot\bm{Y}\rVert_2^2,
\end{equation}
where we set $\alpha=1.0$ in $\bm{W}_{\bm{I}}$ for the basic measures. For the human-centered measures, we set $w_{ij}=1$ for backgrounds and $w_{ij}=5$ for human-regions to accelerating training.

Explicitly optimizing the GLC measure is complicated since it introduces much additional cost including reading and processing a group of photos and color space conversion. To simplify and speed up the training process, we introduce a strategy to simulate the group-level variation using a single image, the pipeline of which is shown in Figure~\ref{GLC_pipeline}. Specifically, given an input $\bm{I}$, we randomly crop two overlapped patches $\bm{I}_{C_1}$ and $\bm{I}_{C_2}$ to mimic the view change in a group of photos. We further randomly adjust the two crops regarding tonal attributes such as temperatures and exposures to synthesize the change of lighting condition and camera setting. We feed the two synthesized crops into a PPR model, obtaining two predictions $\hat{\bm{I}}_{C_1}$ and $\hat{\bm{I}}_{C_2}$ and their overlapped ranges $\hat{\bm{I}}_{C_1}^O$ and $\hat{\bm{I}}_{C_2}^O$. The GLC can be optimized using the following constraint:
\begin{equation}
	\mathcal{L}_{GLC} = \lVert \hat{\bm{I}}_{C_1}^O - \hat{\bm{I}}_{C_2}^O\rVert_2^2.
\end{equation}

The total loss is calculated as 
$
	\mathcal{L} = \mathcal{L}_{HC} + \lambda\mathcal{L}_{GLC},
$
where $\lambda$ is a constant parameter to balance the two losses. We simply set $\lambda=1$ in the experiments.

\section{Experiments}

\subsection{Experiment settings}

\begin{figure*}[t]
		\centering
		\includegraphics[width=1\textwidth]{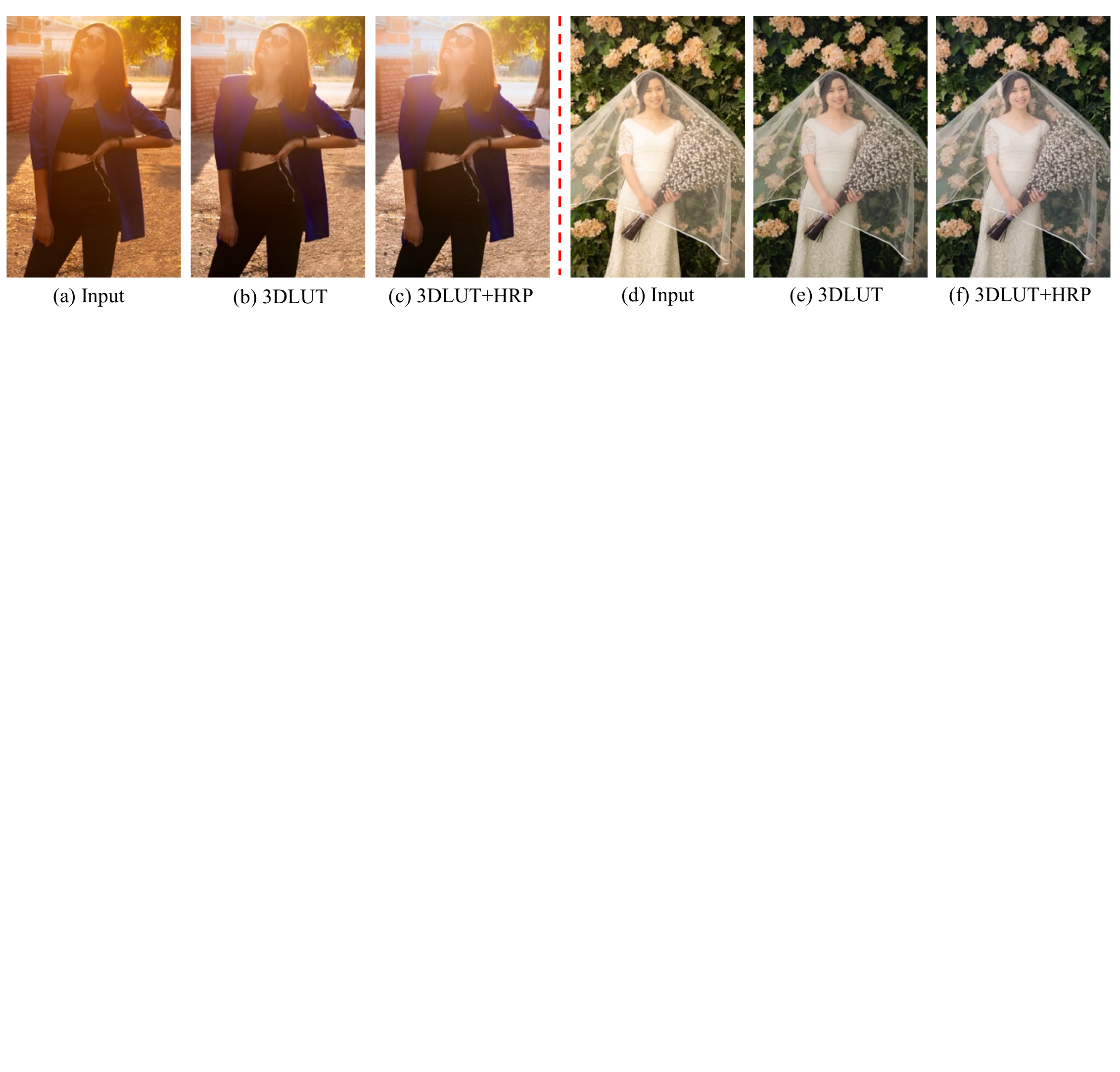}
		\vspace{-35.2em}
		\caption{Visual evaluation of the HRP learning strategy on example photos by using 3D LUT \cite{zeng2020learning}. Leveraging the HRP loss leads to brighter faces and more natural temperature on both examples. \label{comp_HRP}}
\end{figure*}

\noindent\textbf{Datasets:} We employed two datasets, including the constructed PPR10K dataset and the general-purpose FiveK~\cite{bychkovsky2011learning} dataset, in our experiments.
The PPR10K dataset is randomly divided into a training set with $1, 356$ groups and $8, 875$ photos, and a testing set with $325$ groups and $2, 286$ photos. The FiveK dataset is randomly divided into a training set with 4,500 images and a validation set with 500 images following the common practice.
Input images are pre-processed in a 16-bit \textit{tiff} format via CameraRaw to preserve as much information from the raw file as possible, while the target images are converted into 8-bit sRGB color space for convenient display on common devices. To speed up the training process, training images are resized to 360p (short side of the images) resolution. The testing images have two versions: the 360p resolution and the original resolution ranging from 4K to 8K.

\noindent\textbf{Baseline Methods:} Since in practice PPR needs to process very high-resolution photos, this hinders the real applications of most previous photo retouching/enhancement models because of their heavy computational and memory costs. We employ three competitive and efficient models, including the HDRNet~\cite{gharbi2017deep}, the CSRNet~\cite{he2020conditional} and the 3D LUT~\cite{zeng2020learning}, in our experiments (source codes released by authors). To better model such a large-scale and diverse dataset, for the 3D LUT~\cite{zeng2020learning} mothod, we employ $5$ LUTs and use the Resnet-18~\cite{he2016deep} (initialized with the weights pre-trained on ImageNet~\cite{deng2009imagenet}) as the scene classifier.

\noindent\textbf{Data Augmentation:} Besides the commonly used data augmentation methods such as flipping and rotation, we also augment the training images by adjusting $6$ visual attributes in CameraRaw, \ie, temperature, tint, exposure, highlights, contrasts and saturation, to enrich the lighting and color distributions of the training set. The augmentation details can be found in the \textbf{supplementary material}.

\subsection{Baseline Performance}

We first evaluate the baseline performance of the three state-of-the-art photo retouching/enhancement methods on our PPR10K dataset. We retrained each model on each of the three expert sets independently and report their performance under five measures ($PSNR$, $\triangle E_{ab}$, $PSNR^{HC}$, $\triangle E_{ab}^{HC}$, $\mathcal{M}_{GLC}$) in Table \ref{comparison} (rows 1-3, 7-9, 13-15). Each measure is evaluated on two resolutions (360p low resolution (LR) and original high resolution (HR)). Several observations can be made from the results. 

First, all the three models can obtain reasonable results on $PSNR$ and $\triangle E_{ab}$, which indicates the high-quality and self-consistent annotations of the three experts. Among the three versions, the retouching style of expert-a is relatively easier to be learned as expected, since this expert prefers rendering a stronger and stable tonal style for all scenes, leading to a compact target space which is relatively easier to be modeled.
 %to achieve a more catching visual appearance
In contrast, the other two experts prefer a mild rendition to preserve the naturalness of photos (visual examples are provided in the \textbf{supplementary file}). Among the three models, 3D LUT~\cite{zeng2020learning} achieves consistently better performance in most cases than HDRNet~\cite{gharbi2017deep} and CSRNet~\cite{he2020conditional}. Given its high performance and high efficiency, we choose 3D LUT as the baseline model to study the proposed learning strategies in Section \ref{learning strategies}.

\subsection{Models Trained on FiveK and PPR10K}

This section compares the PPR performance of the three methods by training them on the FiveK dataset and on our PPR10K dataset, respectively. We employed the commonly used expert C as the target on the FiveK dataset to train the three models. Input images were processed to have the same format as in our PPR10K dataset. We evaluated the trained models on the three testing sets of PPR10K and report the quantitative results in Table \ref{fivek}. Qualitative comparisons are shown in Figure~\ref{visual_comparison}. As expected, all models trained on the FiveK dataset obtain much worse performance on all measures compared to their counterparts trained on the PPR10K dataset (refer to Table \ref{comparison}), because of the domain gap between general-purpose photo enhancement and PPR. As shown in Figure~\ref{visual_comparison}, the results obtained by the FiveK models have two obvious problems. First, the tone and color appearance of each individual photo is unpleasing especially on the human regions. Specifically, the girl's face is dark in the shadow with unnatural color. Second, the retouched results in a group have large variations on both global tone and local contrast. For example, the third photo has obviously higher brightness and more natural color compared to the first one. In contrast, models trained on our PPR10K dataset achieve not only better individual visual quality but also higher group-level consistency.

\begin{figure*}[t]
		\centering
		\includegraphics[width=1\textwidth]{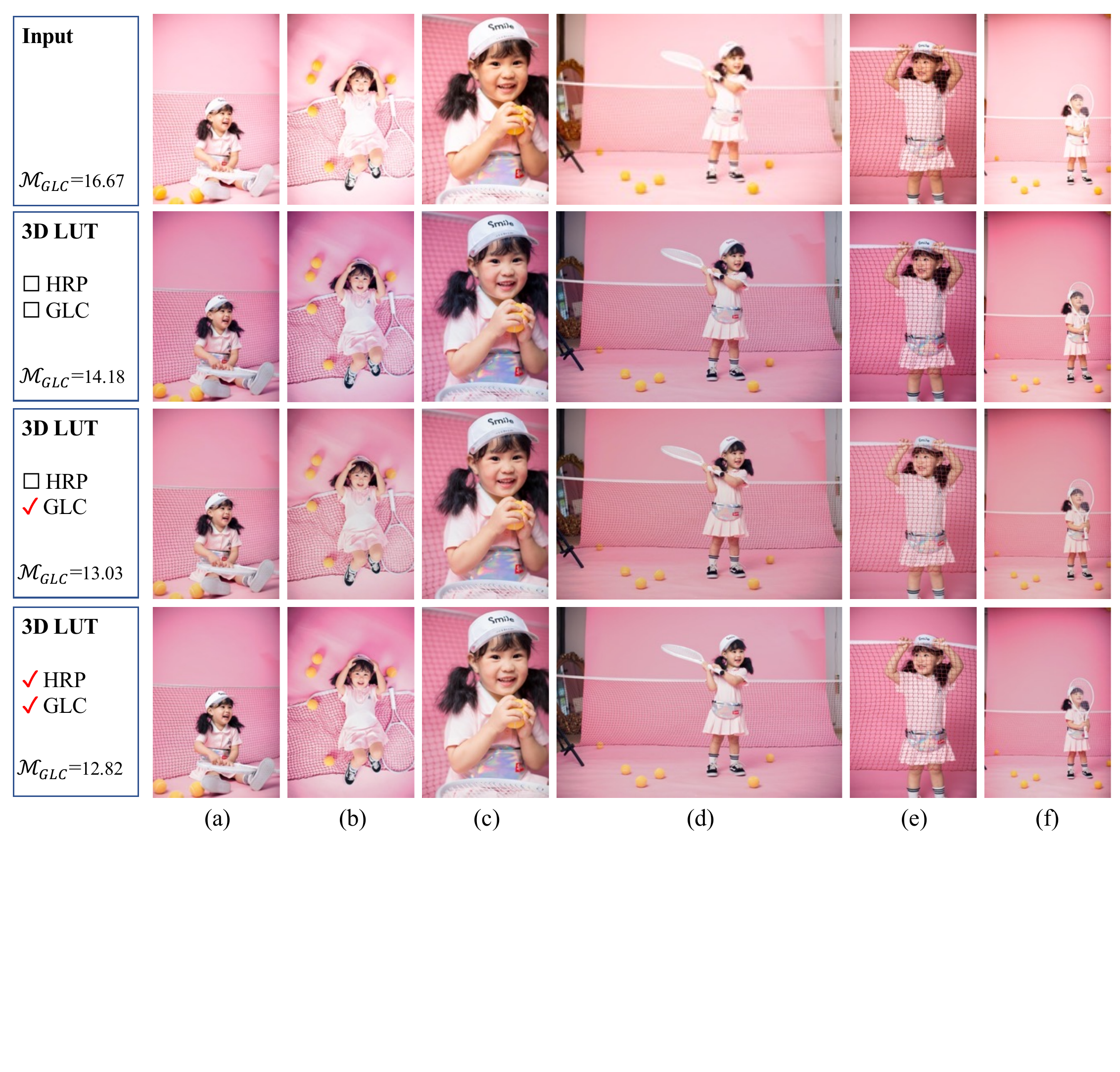}
		\vspace{-11.8em}
		\caption{Visual evaluation of the GLC and GLC+HRP learning strategy by using 3D LUT.
		From \textit{top} to \textit{bottom}: the inputs, results of baseline 3D LUT, results of [3D LUT+GLC] and results of [3D LUT+GLC+HRP].			\label{comp_GLC}}
\end{figure*}

\subsection{Effectiveness of the Learning Strategies}\label{learning strategies}
This section evaluates the effectiveness of the proposed learning strategies using the 3D LUT model. On each of the PPR10K set, we trained three 3D LUT models using only HRP, only GLC, both HRP and GLC learning strategies and report the results in Table \ref{comparison} (rows 4-6, 10-12, 16-18).

One can see that using the HRP loss brings better results on most individual measures. This is reasonable since all three experts paid special attention to the human regions during their retouching. Putting higher weights on human regions thus leads to better individual retouching quality. Two typical visual examples are shown in Figure \ref{comp_HRP}. One can see that using the HRP loss leads to better visual quality (brighter faces and more natural temperature on both examples) on the human regions. 

Using the GLC loss slightly deteriorates the four individual measures but improves the GLC measure. A qualitative example of learning with the GLC loss is shown in Figure \ref{comp_GLC}. As shown in the figure, compared to the results obtained by baseline 3D LUT, the color of the background tends to be more consistent when GLC loss is employed. Specifically, the color of curtain in Figure \ref{comp_GLC} (b, d, f) varies in baseline 3D LUT, while being a consistent pink when the GLC loss is employed.
Another observation is that combining the GLC and HRP losses further improves the GLC measure. This is possibly because jointly optimizing the HRP and GLC losses enables the model to learn complementary information and consequently achieves a good trade-off between individual visual quality and group-level consistency.

\section{Conclusion}

We constructed a large-scale PPR dataset, which was the first of its kind to the best of our knowledge. We collected high quality raw portrait photos with diverse contents from individual photographers and professional photography studios. After careful screening, $11,161$ portrait photos were selected, which fell into $1,681$ groups. High quality human region masks were provided in the dataset. We invited three expert retouchers to label the photos with priority to the human region and the tonal consistency within a group of photos. We defined a set of human-region centered and group-level consistency measures to faithfully evaluate the performance of a PPR model, and accordingly proposed learning strategies to train high quality PPR models. Extensive experiments were conducted to demonstrate the value of the constructed dataset, and the effectiveness of the proposed measures and learning strategies.

{\small
\bibliographystyle{ieee_fullname}
\bibliography{bib_CVPR2021}
}

\end{document}